\documentclass[runningheads]{llncs}
\usepackage[T1]{fontenc}
\usepackage{graphicx}
\usepackage{comment, algorithm, algpseudocode, amsmath}
\usepackage{pgfplots, subcaption, tikz, svg}
\pgfplotsset{compat=1.18}
\usepackage{xcolor}

\begin{document}
\title{Exploring the Design Space of Fair Tree Learning Algorithms}
%
%
\author{Kiara Stempel\orcidID{0009-0001-9978-7872} \and
Mattia Cerrato\orcidID{0000-0001-7736-0547} \and
Stefan Kramer\orcidID{0000-0003-0136-2540}}
\authorrunning{K. Stempel \and M. Cerrato \and S. Kramer}
%
\institute{
Johannes Gutenberg University Mainz, Staudingerweg 9, 55128 Mainz, Germany
\email{stempel@uni-mainz.de} \\
}
\maketitle 
\begin{abstract}
Decision trees have been studied extensively in the context of fairness, aiming to maximize prediction performance while ensuring non-discrimination against different groups. Techniques in this space usually focus on imposing constraints at training time, constraining the search space so that solutions which display unacceptable values of relevant metrics are not considered, discarded, or discouraged.
If we assume one target variable $y$ and one sensitive attribute $s$, the design space of tree learning algorithms can be spanned as follows: 
(i) One can have {\em one} tree $T$ that is built using an objective function that is a function of $y$, $s$, and $T$. For instance, one can build a tree based on the weighted information gain regarding $y$ (maximizing) and $s$ (minimizing). 
(ii) The second option is to have {\em one} tree model $T$ that uses an objective function in $y$ and $T$ and a constraint on $s$ and $T$. Here, $s$ is no longer part of the objective, but part of a constraint. This can be achieved greedily by aborting a further split in a subspace as soon as the splitting condition that optimizes the objective function in $y$ does not satisfy the constraint on $s$. The simplest form to examine other possible splits in this subspace is to backtrack in tree construction, once a fairness constraint in terms of $s$ is violated in a subspace of a new split. 
(iii) The third option is to have {\em two} trees $T_y$ and $T_s$, one for $y$ and one for $s$, such that the tree structure for $y$ and $s$ does not have to be shared. In this way, information regarding $y$ and regarding $s$ can be used independently, without having to constrain the choices in tree construction by the mutual information between the two variables. Quite surprisingly, of the three options, only the first one and the greedy variant of the second have been studied in the literature so far. In this paper, we introduce the above two additional options from that design space and characterize them experimentally on multiple datasets.

\end{abstract}
\section{Introduction}
\label{sec:intro}

As machine learning models increasingly influence critical decision-making processes, addressing fairness has become an essential field of research. Algorithmic fairness involves ensuring that model predictions do not disadvantage individuals based on protected groups such as religion, gender, or race.
Among the various algorithms designed or adapted to mitigate discrimination, decision trees belong to the more transparent methods,
as they provide a clear and interpretable structure for addressing fairness concerns. 
Various attempts have been made to incorporate fairness into decision trees, typically focusing on imposing constraints during training, e.g., constraining the search space to discard, discourage, or exclude solutions that lead to unacceptable values of relevant fairness metrics by adjusting the splitting criterion. 

To account for both, performance based on the target variable $y$, and fairness based on the sensitive attribute $s$, the design space allows for (i) constructing \textit{one} tree $T$ by combining performance and fairness in a single objective function, for instance by weighting the information gain regarding $y$ and $s$, (ii) constructing \textit{one} tree $T$ by imposing a fairness constraint on the objective function, or (iii) modeling performance and fairness independently in \textit{two} separate trees $T_y$ and $T_s$.
While prior work has mostly focused on the first and, to a limited extent, the second approach, the third design option remains unexplored. However, it allows for an independent use of information regarding $y$ and $s$, and provides the possibility to examine predictions and decision paths of the two separate trees.

In this paper, we provide a broader perspective on fair tree learning. We experimentally evaluate the above design choices for fair tree learning and identify their characteristics. 
For this purpose, we introduce a two-tree approach that separates fairness and performance into distinct learning objectives and therefore falls into (iii). Further, we propose a backtracking mechanism for (ii) that actively revises previous splitting decisions when fairness constraints are at risk of being violated.
The evaluation is based on the trade-off curves of the tree variants. 
We obtain a trade-off curve by varying a parameter $\gamma$ that controls the trade-off between performance and fairness.
Several measures derived from the trade-off curve are used to compare the approaches.
The area under the trade-off curve (AUTOC), along with the number of unique and local Pareto-optimal points, provides a compact summary of the trade-off curves, while offering different perspectives for evaluation.


\section{Related Work} 
\label{sec:related_work}

Several methods have been proposed to adapt decision trees to ensure fairness while maintaining predictive performance, typically by modifying the splitting criterion to incorporate fairness considerations:
Kamiran et al.~\cite{kamiran_discrimination_2010} introduced the first discrimination-aware decision trees by modifying the splitting criterion to balance information gain on the target and sensitive attribute, combining them via subtraction, addition, or division. The method penalizes splits that reveal too much about sensitive groups and is complemented by post-hoc leaf relabeling.
This idea was extended to fair forests by Raff et~al.~\cite{raff_2020}, building fair decision trees similar to the ones by Kamiran~et~al. The difference between the Gini impurity~\cite{CART} on the class label and the information gain of the protected attribute seeks to balance predictive performance with fairness.
Castelnovo et al.~\cite{castelnovo_fftree_2020} modify the splitting criterion to do a constrained search for the best attribute. They maximize the information gain on $y$ while enforcing a threshold for a given fairness criterion. 
Since most fair tree learning methods optimize for a fixed threshold on the fairness metric, Barata~et~al.~\cite{barata_2021} propose a splitting criterion based on AUROC to be independent of the decision threshold.

For online learning, Zhang and Ntoutsi~\cite{zhang_faht_2021} extend Hoeffding trees~\cite{Hoeffding} with a fairness gain, which is defined in terms of the reduction in discrimination caused by a specific split.
Zhang and Zhao~\cite{zhang_online_2020} introduce an adaptive fair information gain to Hoeffding trees. Instead of addition, they multiply the predictive and sensitive gains to account for the possible difference in scales between the gains.

An approach to learning optimal fair decision trees
is done by Aghaei et al.~\cite{aghaei_learning_2021}. 
This exact approach unifies the fairness definitions across classification and regression and uses a mixed-integer optimization framework for learning. Their framework incorporates a discrimination regularizer and a corresponding regularization weight that allows tuning the fairness-performance trade-off.

Instead of learning a new, fair model, Zhang et al. suggest repairing an existing unfair tree by flipping or refining specific decision paths in the tree while setting hard constraints such as fairness and semantic difference~\cite{zhang_fair_2020}.

Fair classification can also be framed as a multi-objective optimization problem, balancing performance and fairness. Optimization methods fall into exact and meta-heuristic categories, the latter split into single- and multi-objective techniques~\cite{sharma_2022}. Unlike single-objective optimization, multi-objective optimization seeks a set of Pareto-optimal solutions rather than a single optimum~\cite{Bandyopadhyay_2013}.
Multi-objective evolutionary algorithms use performance indicators to guide search toward Pareto fronts without computing Pareto dominance relations~\cite{sharma_2022}. Zitzler et al. introduced an indicator-based evolutionary algorithm compatible with various indicators~\cite{Zitzler_2004}. The hypervolume indicator transforms the problem into maximizing a single metric~\cite{hypervolumes_auger}. Many approaches aim to improve this indicator, to quantify the volume of the region dominated by a set of solutions.~\cite{hypervolume_auger2010,hypervolume_beume}.
La Cava~\cite{hypervolume_cava} applies the hypervolume to optimize the fairness-performance trade-off via a meta-model that maps protected attributes to sample weights. Freitas and Brookhouse~\cite{freitas_2024} review such multi-objective evolutionary methods in fairness-aware learning.


\section{Methods}
\label{sec:method}

\subsection{Defining the Design Space}
\label{sec:definitions}

In this paper, we address binary classification settings. With $n$ denoting the number of samples and $m$ denoting the number of features, the input variables are represented by matrices $X_{n \times m}$. The corresponding set of class labels is defined as $Y$, where $y_i \in \{0, 1\}$ for each instance $i=1, ..., n$. When training a fair decision tree, the objective is to reduce information about the sensitive attribute $S$, where $s_i \in \{0, 1\}$, to mitigate bias or discrimination based on $S$. 
Let $\mathcal{T}$ be the search space of our classification problem, i.e., the set of all potential decision trees that could be constructed based on $X$, that is
\begin{align}
    \mathcal{T}(X) = \{ T \mid T \text{ is a decision tree based on } X \}
\end{align}
Each point in $\mathcal{T}$ corresponds to a unique tree structure. The objective of a standard decision tree algorithm~\cite{CART} is to find a decision tree $T \in \mathcal{T}$ that maximizes the predictive performance on $X$.
The decision space $\mathcal{D}(T)$ refers to the set of possible outcomes a tree $T$ can produce given any input. That is, $\mathcal{D}(T)$ represents all possible class labels that the fixed tree structure is capable of making.
The set of predictions obtained by applying $T$ to input data is denoted by $\hat{Y}$. 

Based on this foundation, we can now span the design space of fair tree learning algorithms. When learning a fair tree on the input $X$, we are given one target variable $y$ and one sensitive attribute $s$, and the objective is (i) to obtain a good predictive performance while (ii) minimizing the discrimination, or the information about $s$ used for predicting.

One option is to have \textit{one} tree that is built using a single objective function representing both objectives, which in this way is a function of $y$, $s$, and $T$. A typical approach is to adjust the splitting criterion to account for both $y$ and $s$. 
The splitting criterion that is optimized during the training process typically corresponds to the information gain $G$:
\begin{align}
    G_V(A) = H_V(X) - \sum_{a \in \text{Values}(A)} \frac{|X_a|}{|X|} H_V(X_a).
\end{align}
It measures the increase in purity regarding a potential splitting attribute $A$ and target attribute $V$ (which can be either $y$ or $s$), of which we measure the impurity. The impurity of $V$ is given by the entropy $H_V(X)$ of dataset $X$.
Under this design, fairness constraints are integrated into the search space $\mathcal{T}$, since we aim to find the tree model $T$ that optimizes a combined objective function.
For accounting for both performance and fairness, one can have two information gains which are linearly combined, where the information gain on $y$ is maximized and the information gain on $s$ is minimized. 

The second design option, also defined over $\mathcal{T}$, learns \textit{one} tree using an objective based solely on $y$ and $T$, with fairness imposed as a constraint on $s$ and $T$. Instead of weighting fairness in the objective, the trade-off is controlled via thresholds that are part of the constraint.
This can be implemented greedily by aborting further splits in a subspace as soon as the split that best optimizes the objective in $y$ violates the constraint on $s$. A simple extension of this idea is to backtrack during tree construction -- i.e., revisiting earlier splitting decisions -- when a fairness constraint is violated within a subspace.
Beyond the approach of integrating both objectives into a (constrained) single objective function, a third,  alternative design involves separating them into two objective functions. This results in learning \textit{two} separate decision trees: one ($T_y$) optimized for the objective function in $y$ and $T$, and the other ($T_s$) for the objective function in $s$ and $T$. The resulting trees are then combined. In this design, the impact of fairness constraints is effectively shifted from the search space into the decision space $\mathcal{D}(T)$.

\subsection{Extending the Constrained Single-Tree Approach with Backtracking}
\label{sec:backtracking}

To allow for the examination of more possible splits when having one constrained objective function as design variant (ii) and imposing the fairness constraints in search space $\mathcal{T}$, we introduce DTFC (decision tree with fairness constraint), a variant using a backtracking mechanism that adaptively refines the tree construction process. 

\begin{algorithm}[t!]
\caption{\texttt{DTFC}}
\label{algo:backtracking}
\begin{algorithmic}[1]
\State \textbf{Input:} $X \neq \emptyset $, $Y$, $S$, $\gamma$, $max\_depth$, $min\_samples$, current depth $d$
\State \textbf{Output:} A fair decision tree node or \textbf{null} if no valid split found

\State $A \gets$ set of available attributes
\State $A_{\text{valid}} \gets \emptyset$ \Comment{Attributes satisfying the fairness constraint}

\If{A is empty}
    \State \Return Leaf with majority label in $X$
\EndIf

\ForAll{$A_i \in A$}
    \State Compute $\text{G}_Y(A_i)$ and $\text{G}_S(A_i)$
    \If{$\text{G}_S(A_i) \leq \gamma$}
        \State $A_{\text{valid}} \gets A_{\text{valid}} \cup \{A_i\}$
    \EndIf
\EndFor

\State Sort $A_{\text{valid}}$ by decreasing $\text{G}_Y$
\ForAll{$A^* \in A_{\text{valid}}$}
    \State Create node and candidate node splitting on $A^*$
    \State \texttt{valid\_split} $\gets$ \textbf{true}
    \ForAll{values $a$ of $A^*$}
        \State $X_a \gets \{x \in X \mid x[A^*] = a\}$
        \If {$|X_a| \geq \texttt{min\_samples}$ and $d$ < $max\_depth$}
            \State $T_a$ = \texttt{DTFC}($X_a$, $Y$, $S$, $\gamma$, $max\_depth$, $min\_samples$, $d+1$)
            \If{$T_a = \textbf{null}$}
                \State \texttt{valid\_split} $\gets$ \textbf{false}
                \State \textbf{break}
            \EndIf
            \State Attach $T_a$ to current node under $A^* = a$
        \Else \State \Return Leaf with majority label in $X$
        \EndIf        
    \EndFor
    \If{\texttt{valid\_split}}
        \State \Return current node
    \EndIf
\EndFor

\State \Return \textbf{null} \Comment{No fair and valid split found, i.e., backtrack}

\end{algorithmic}
\end{algorithm}

The core idea of the backtracking approach is to allow the decision tree algorithm to revisit and actively refine previous splitting decisions whenever the fairness threshold is compromised. 
In standard fair tree algorithms, splits in tree $T$ are chosen to balance the objective in $y$ and $T$ with the fairness constraint on $s$ and $T$. However, this approach may become suboptimal as deeper splits could amplify bias introduced by earlier decisions. DTFC accepts a candidate split only if the information gain on the sensitive attribute $s$ remains below a threshold $\gamma$; otherwise, alternative attributes are considered. If no valid split is found, the algorithm backtracks to explore other options at earlier nodes, allowing the algorithm to revisit and correct biased splits. Algorithm \ref{algo:backtracking} provides a formal description of the procedure.
The backtracking mechanism of DTFC enables deeper trees by exploring more splitting options than the constrained approach without backtracking. Rather than creating a leaf when fairness thresholds are violated, it searches for alternative splits that may better balance performance and fairness. If no valid configuration is found, DTFC returns an empty tree, ensuring that fairness is not compromised.

However, the backtracking approach is not without limitations. The primary drawback is its computational inefficiency, as the process of revisiting and evaluating alternative splits can introduce a substantial computational overhead compared to conventional tree-growing methods. Although backtracking allows for local adjustments at individual nodes, it does not necessarily optimize the tree globally. Decisions are still made on a greedy node-by-node basis, which may prevent the algorithm from achieving the overall optimal structure for performance and fairness trade-offs. 

\subsection{Dual-Tree Approach}

The dual-tree approach, 2TFT (Dual Trees for Fairness Trade-Offs), separates the training process of fair decision trees into two distinct trees, one optimizing the performance-related objective function, the other optimizing the fairness-related objective function, to shift the trade-off between predictive performance and fairness constraints to $\mathcal{D}(\mathcal{T})$. As $T_y$, we will denote a tree that is exclusively performance-aware and thus has a fully performance-related objective in $y$ and $T_y$, whereas $T_s$ refers to a tree that is only fairness-aware and has a fully fairness-related objective in $s$ and $T_s$.
In the training procedure, $T_y$ is trained as a standard decision tree, i.e. the splitting criterion optimizes only the information gain on the target variable, $G_Y$. In this way, $T_y$ does not address any fairness metrics and is therefore fairness-unaware.

\begin{figure}[t]
    \centering
    \includegraphics[width=0.95\textwidth]{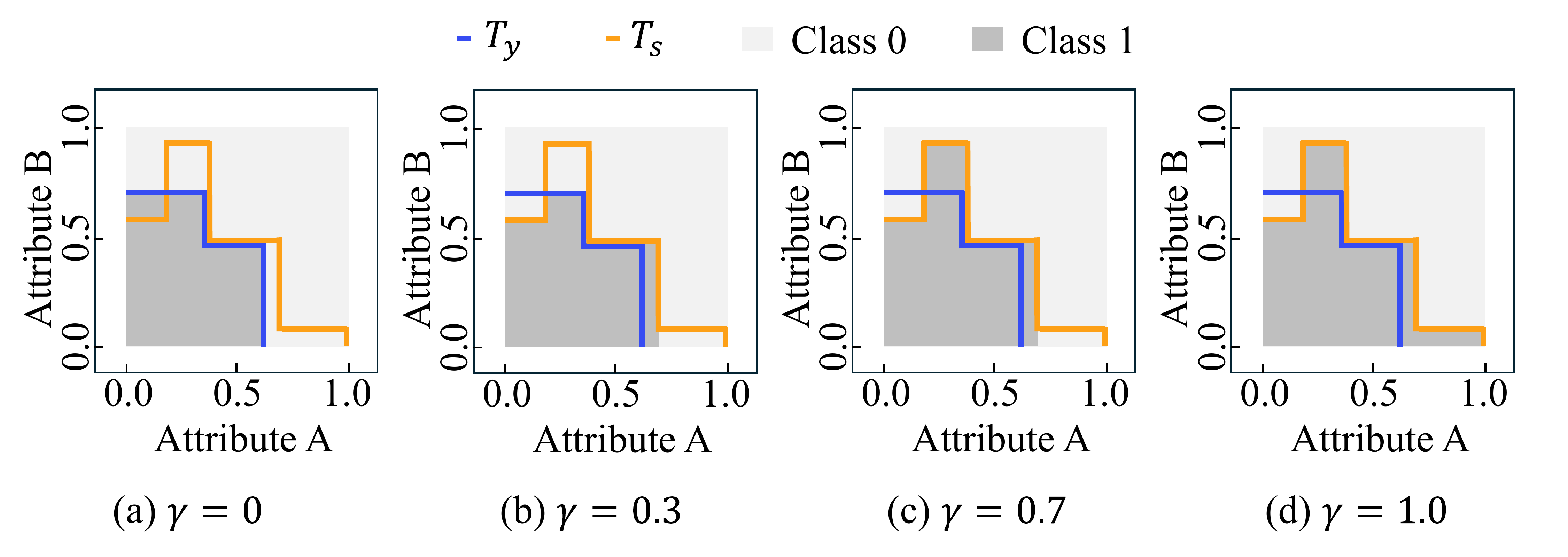}
    \caption{Model partitions of 2TFT for varying $\gamma$ in the case of two features. Partitions are colored by predicted class, depending on whether $T_y$ or $T_s$ dominates.}
    \label{fig:visualization}
\end{figure}

The splitting criterion of $T_s$ is adjusted similarly to the one described in Section \ref{sec:related_work}. We want $T_s$ to be fairness-aware, but performance-unaware. Therefore, we do not maximize the information gain on $y$ as we do for $T_y$. Rather, in each splitting step, we choose the attribute minimizing the information gain $G_S$ we obtain on $S$, i.e.,
we maximize $- G_S$. However, also with this tree, we predict with respect to $y$. The class probability estimates are calculcated as the proportion of samples in a leaf that belong to a certain class.
After training, 2TFT combines the class probability estimates of $T_y$ and $T_s$ in $\mathcal{D}(\mathcal{T})$ in a linear combination
\begin{align}
    T(x) = (1-\gamma) \cdot T_y(x) + \gamma \cdot T_s(x),
\end{align}
where $\gamma$ is the parameter that allows for controlling the trade-off. Figure \ref{fig:visualization} visualizes the combined predictions, resulting in different partitions of $\mathcal{D}(\mathcal{T})$.

In the process of making the tree model fair, two separate trees allow for more flexibility regarding the structure of the decision trees, since their structures can be defined and optimized independently of each other. 

\subsection{Comparison Methods}

The fair tree learning methods implemented for comparison correspond to the following design options: a combined objective optimizing both performance and fairness in one tree for design variant (i) (as introduced by Kamiran et al. \cite{kamiran_discrimination_2010}), and a constrained objective enforcing threshold on $G_S$ during splitting for variant (ii) (as described by Castelnovo et al. \cite{castelnovo_fftree_2020}).

The combined-objective approach adjusts the splitting criterion by adding the information gain on $Y$ ($G_Y$) and the negative gain on $S$ ($G_S$), weighted by a trade-off parameter $\gamma$:
$\texttt{max} (1-\gamma) \cdot G_Y - \gamma \cdot G_S.$
The constrained approach maximizes $G_Y$ subject to $G_S \leq \gamma$. If no attribute satisfies this constraint, a leaf node is created.
By varying $\gamma$, we obtain a list of different models representing different possible fairness-performance trade-offs. These trade-offs can subsequently be expressed as sequences of values of performance and fairness measures.


\section{Experimental Evaluation of the Trade-Off Curves}
\label{sec:experiments}

The experiments are designed to assess performance–fairness trade-offs across methods spanning the design space of fair tree learning. 
This section introduces the fairness and performance metrics used to construct trade-off curves, the evaluation criteria for analyzing them, and the experimental setup, including datasets, baselines, and hyperparameters.
We also discuss interpretability and the runtime of the methods. Our Python implementation is openly available\footnote{https://github.com/kiarastempel/FairTreeExperiments}.

\subsection{Evaluation Measures}
\label{sec:measures}

We measure the performance of a decision tree $T$ in $\mathcal{D(T)}$ using AUROC. The strength of discrimination can be measured using several different fairness metrics. We focus on the statistical parity difference~\cite{verma_rubin}, defined as
\begin{align}
    \text{SPD} = \left| P(\hat{Y} = 1 \mid S = 0) - P(\hat{Y} = 1 \mid S = 1) \right|.
\end{align}
The trade-off curve is created with AUROC on the x-axis and $1-\text{SPD}$ on the y-axis. 
Based on this, we outline the following measures for analyzing the curve.
\paragraph{AUTOC} 
Computed as the area under the trade-off curve, AUTOC captures the overall balance between fairness and performance, aiming for points near the upper-right corner of the performance–fairness trade-off space. 
The curve is represented by a set of points obtained by varying the $\gamma$ parameter, encoded in two lists: \texttt{SPD} (fairness values) and \texttt{AUROC} (corresponding performance values).
All curves are extended by two points, where we add one point to $(0.5, \text{max}(1-\text{SPD}))$ and another one to $(1, \text{min}(1-\text{SPD}))$, to avoid penalizing steep curves with consistently high AUROC values but limited AUROC range. 
All points are sorted increasingly by AUROC before approximating the area under the curve using the trapezoidal rule \cite{DahlquistBjorck2008}. Consequently, we formally define AUTOC as
\begin{align}
\begin{split}
    \text{AUTOC} & = \frac{1}{2} \Big[\big(\texttt{AUROC}_0 - 0.5\big)  \big(\text{max}(1-\texttt{SPD}) + (1-\texttt{SPD}_0)\big) \\
    & \quad + \sum_{k=0}^{K} \big(\texttt{AUROC}_{k+1} - \texttt{AUROC}_k\big) \big((1-\texttt{SPD}_k) + (1-\texttt{SPD}_{k+1})\big) \\
    & \quad + \big(1 - \texttt{AUROC}_K\big)  \big((1-\texttt{SPD}_K) + \text{min}(1-\texttt{SPD})\big)\Big]. \\
\end{split}
\end{align}
The measure is motivated by the definition of hypervolumes, as already mentioned in Section \ref{sec:related_work}, to evaluate the quality of a set of solutions in multi-objective optimization, particularly in terms of their proximity to the Pareto front, i.e., the optimal trade-offs~\cite{hypervolume_beume,hypervolumes_auger}. 
If we map each solution $T \in \mathcal{D}(\mathcal{T})$ to a point in the objective space $\mathcal{F}$, which represents the set of all possible outcomes of the objective functions evaluated on the models in $\mathcal{T}(X)$, then the hypervolume quantifies the volume of the region dominated by a set of solutions in $\mathcal{F}$.

\paragraph{Number of Points on Pareto Front}
The Pareto front represents the set of solutions where no single objective can be improved without worsening the other~\cite{Kang2024}. 
In this context, a point is Pareto-optimal within one trade-off curve (we call this locally Pareto-optimal) if no other point in the trade-off curve achieves both higher fairness and better performance.
The number of such points reflects the model's ability to generate optimal trade-offs, with higher counts indicating that the model provides users with more flexibility to prioritize fairness or performance as needed. 
Ideally, all trade-off points are Pareto-optimal, but this may be reduced by redundancy.

\paragraph{Number of Unique Points}

This measure counts distinct trade-off combinations, considering identical AUROC and SPD values as duplicates. It reflects how solution diversity evolves under varying fairness constraints, indicating whether changes in the strength of a fairness constraint lead to gradual, smooth performance shifts or sudden, significant impacts.

\paragraph{Variance of Pairwise Distances}
The distribution variance evaluates how evenly the points on the trade-off curve are distributed across the fairness-performance space. 
It is computed as the variance of pairwise Euclidean distances between all points on the curve, with a value of 0 for trade-off curves with a single unique point. 
Low variance may indicate that the curve is concentrated in specific regions or clusters, while a higher variance suggests a more uniform coverage of the decision space, which makes this measure relevant when assessing the practical usability of the curve and corresponding models.

\subsection{Experimental Setup}

For each dataset, the class labels $Y$ and sensitive groups $S$ are transformed to being either 0 or 1. Further, all categorical features are one-hot encoded.
When training a single tree, for each numerical attribute, 10 equally distributed thresholds are tested for a split.
For evaluation, as recommended by Claude and Bengio~\cite{claude_bengio}, we perform a 15-times hold-out in the outer loop, splitting the data randomly into $\frac{2}{3}$ of training data and $\frac{1}{3}$ of test data for each hold-out. 
In each hold-out, the hyperparameters defining the tree size are optimized on the training data using three-times cross-validation in an inner loop. 
We evaluate all hyperparameter combinations on the three folds by calculating the trade-off curve for all of them. Then, we choose the one where the averaged AUTOC is maximized, and compute all measures as described above. 
The best found hyperparameters are selected for retraining and evaluating the model on the held-out dataset.

When creating the trade-off curve for a specific hyperparameter combination, the interval of $\gamma$ is set to $[0, 1]$ for those approaches with weighted combinations or soft constraints (i.e., one tree with combined splitting criterion, and two trees), while $[0, 0.2]$ is used for the approaches using hard fairness-related constraints (i.e., one tree with constrained splitting criterion, the backtracking version, and the relaxed threshold optimizer) to account for the smaller range of the information gain.
We test 50 equally distributed values of $\gamma$ for each run that creates a trade-off curve.
To ensure consistency, all models were trained under identical hyperparameter settings. The selected hyperparameters are shown in Table \ref{tab:hps}. The value of $\texttt{min\_samples}$ refers to the proportion of instances of the dataset that a leaf node has to obtain at minimum, i.e., if at least one potential child node of the current node gets less than  $n \cdot$ \texttt{min\_samples} training instances, then the current node becomes a leaf. The maximum depth a tree can reach in the training process is set with $\texttt{max\_depth}$.
\begin{table*}[t]
\caption{Tested hyperparameter combinations for all methods based on trees. For backtracking, we report all hyperparameter combinations that were tested within the time range of 10 hours. The value of \texttt{min\_samples} corresponds to the proportion of instances in the given dataset.}
    \centering
    \setlength{\tabcolsep}{1mm}
    \resizebox{0.99\textwidth}{!}{%
    \begin{tabular}{lll}
        \hline
        & Dataset & Tested combinations of (\texttt{max\_depth}, \texttt{min\_samples}) \\ 
        \hline
        General          & & All combinations with \\
                         & & \quad - \texttt{max\_depth} = \{4, 6, 8, 13\} \\
                         & & \quad - \texttt{min\_samples} = \{0.25, 0.1, 0.01\} \\
        DTFC     & COMPAS           & (4, 0.25), (4, 0.1), (4, 0.01), (6, 0.25), (6, 0.1), (6, 0.01), (8, 0.25), (8, 0.1) \\
        & Adult            & (4, 0.25), (4, 0.1), (4, 0.01), (6, 0.25), (6, 0.1) \\
        & Dutch            & (4, 0.25), (4, 0.1), (4, 0.01), (6, 0.25), (6, 0.1) \\
        \hline
    \end{tabular}
    }
    \label{tab:hps}
\end{table*}
Since the backtracking approach is by far the slowest, especially for deeper tree sizes, we set a time limit of 10 hours per dataset and evaluate the best hyperparameters out of all combinations that were tested in that time range.
After choosing the 15 best trade-off curves based on AUTOC, we evaluate the found curves subsequently on the other criteria described in Section \ref{sec:measures}.

\subsection{Datasets}
\label{sec:datasets}

We apply the described methods to the COMPAS dataset, 
the Adult dataset, and the Dutch Census dataset from 2001 \cite{fairness_datasets}.

The COMPAS dataset is widely used to predict criminal recidivism and includes features like prior convictions, age, gender, and race. In this context, race is considered as sensitive attribute $S$.
Another widely studied dataset is the Adult dataset, derived from U.S. Census data, and is commonly used to predict whether an individual earns over~\$50K annually. It includes features such as age, education, occupation, race, and gender, the latter typically serving as the sensitive attribute $S$, highlighting potential gender bias in income prediction.
Similarly, the Dutch Census dataset, based on the data from 2001, contains demographic information and is, like Adult, commonly used to explore gender-based discrimination, also in these experiments.

\subsection{Baseline}
\label{sec:baseline}

Our baseline is the threshold optimizer, an implementation by fairlearn~\cite{bird2020fairlearn}, based on work by Hardt et al.~\cite{Hardt2016}. 
Unlike tree-based in-processing methods that incorporate fairness constraints during training, this is a post-processing algorithm that adjusts prediction thresholds after training to enforce fairness.
It modifies decision thresholds separately for each sensitive group to exactly satisfy a fairness criterion without altering the underlying model. However, in contrast to in-processing methods, it requires sensitive attributes to be available at prediction time. 
Since this post-adjustment of the threshold is optimal, we expect not to be better than postprocessing. 
In our implementation, we used a forked version of the repository\footnote{https://github.com/taharallouche/fairlearn}, which includes a relaxed variant of the threshold optimizer, but is open for merge with the original fairlearn library. Here, a tolerance value for the given fairness metric can be specified to create a trade-off curve. As a base model, we pass a decision tree classifier.


\subsection{Results and Discussion}

In this section, we discuss our experimental results on the datasets described in Section \ref{sec:datasets} and the methods introduced in Section \ref{sec:method}.



Analyzing the trade-off curves of the implemented tree variants provides insights into their ability to balance performance and fairness objectives. The primary focus is to assess how different decision tree models perform under varying constraints, represented by the trade-off parameter $\gamma$. This evaluation aims to highlight the strengths and weaknesses of the different approaches and their potential applicability in real-world scenarios.

\begin{figure*}[t]
    \centering
    \begin{minipage}[t]{0.9\textwidth}
    \ref{combinedlegend}
    \end{minipage}
    
    \begin{minipage}[t]{0.33\textwidth}
        \centering
        \begin{tikzpicture}[scale=0.57]
            \begin{axis}[
                scale=0.8, 
                ylabel={1 - SPD},
                xlabel={AUROC},
                xmin=0.5,
                ymax=1,
                grid=both,
                major grid style={line width=0.1pt, draw=gray!50, opacity=0.3},
                minor grid style={line width=0.1pt, draw=gray!30, opacity=0.3},
                ylabel style={yshift=-2pt},
                label style={font=\Large },
                tick label style={font=\large},
                legend style={ 
                    at={(0.5,10.3)}, 
                    anchor=north, 
                    draw=none,
                    legend columns=2,
                    font=\footnotesize, 
                }, 
                legend to name=combinedlegend
            ]

            \addplot[
                scatter src=explicit,
                only marks,
                mark=o, 
                mark size=2.0pt,
                color=green,
                table/col sep=comma,
                ]
                table [
                y={1-spds_test}, 
                x={aurocs_test}
            ] {results/avg_curve_Compas_two_trees.csv};
            \addlegendentry{2TFT (Two trees)}

            \addplot[
                scatter src=explicit,
                only marks,
                mark=triangle,
                mark size=2.0pt,
                color=red,
                table/col sep=comma,
                ]
                table [
                y={1-spds_test}, 
                x={aurocs_test}
            ] {results/avg_curve_Compas_weighted_combi.csv};
            \addlegendentry{One tree (combined split criterion)}

            \addplot[
                scatter src=explicit,
                only marks,
                mark=square,
                mark size=2.0pt,
                color=blue,
                table/col sep=comma,
                ]
                table [
                y={1-spds_test}, 
                x={aurocs_test}
            ] {results/avg_curve_Compas_constrained.csv};
            \addlegendentry{One tree (constrained split criterion)}

            \addplot[
                scatter src=explicit,
                only marks,
                mark=diamond,
                mark size=2.0pt,
                color=brown,
                table/col sep=comma,
                ]
                table [
                y={1-spds_test}, 
                x={aurocs_test}
            ] {results/avg_curve_Compas_backtracking.csv};
            \addlegendentry{DTFC (backtracking)}

            \addplot[
                scatter src=explicit,
                only marks,
                mark=+,
                mark size=2.0pt,
                color=darkgray,
                table/col sep=comma,
                ]
                table [
                y={1-spds_test}, 
                x={aurocs_test}
            ] {results/avg_curve_Compas_postprocessing.csv};
            \addlegendentry{Threshold Optimizer}

            \end{axis}
        \end{tikzpicture}
        \subcaption{Compas}
    \end{minipage}
    \begin{minipage}[t]{0.32\textwidth}
        \centering
        \begin{tikzpicture}[scale=0.57]
            \begin{axis}[
                scale=0.8, 
                ylabel={1 - SPD},
                xlabel={AUROC},
                xmin=0.5,
                ymax=1,
                grid=both,
                major grid style={line width=0.1pt, draw=gray!50, opacity=0.3},
                minor grid style={line width=0.1pt, draw=gray!30, opacity=0.3},
                ylabel style={yshift=-2pt},
                label style={font=\Large },
                tick label style={font=\large},
                legend style={ 
                    at={(0.5,10.3)}, 
                    anchor=north, 
                    draw=none,
                    legend columns=2,
                    font=\footnotesize, 
                }, 
                legend to name=combinedlegend
            ]

            \addplot[
                scatter src=explicit,
                only marks,
                mark=o,
                mark size=2.0pt,
                color=green,
                table/col sep=comma,
                ]
                table [
                y={1-spds_test}, 
                x={aurocs_test}
            ] {results/avg_curve_Adult_two_trees.csv};
            \addlegendentry{2TFT (Two trees)}

             \addplot[
                scatter src=explicit,
                only marks,
                mark=triangle,
                mark size=2.0pt,
                color=red,
                table/col sep=comma,
                ]
                table [
                y={1-spds_test}, 
                x={aurocs_test}
            ] {results/avg_curve_Adult_weighted_combi.csv};
            \addlegendentry{One tree (combined split criterion)}

            \addplot[
                scatter src=explicit,
                only marks,
                mark=square,
                mark size=2.0pt,
                color=blue,
                table/col sep=comma,
                ]
                table [
                y={1-spds_test}, 
                x={aurocs_test}
            ] {results/avg_curve_Adult_constrained.csv};
            \addlegendentry{One tree (constrained split criterion)}

            \addplot[
                scatter src=explicit,
                only marks,
                mark=diamond,
                mark size=2.0pt,
                color=brown,
                table/col sep=comma,
                ]
                table [
                y={1-spds_test}, 
                x={aurocs_test}
            ] {results/avg_curve_Adult_backtracking.csv};
            \addlegendentry{DTFC (backtracking)}
            
            \addplot[
                scatter src=explicit,
                only marks,
                mark=+,
                mark size=2.0pt,
                color=darkgray,
                table/col sep=comma,
                ]
                table [
                y={1-spds_test}, 
                x={aurocs_test}
            ] {results/avg_curve_Adult_postprocessing.csv};
            \addlegendentry{Threshold Optimizer}
            
            \end{axis}
        \end{tikzpicture}
        \subcaption{Adult}
    \end{minipage}
    \begin{minipage}[t]{0.32\textwidth}
        \centering
        \begin{tikzpicture}[scale=0.57]
            \begin{axis}[
                scale=0.8, 
                ylabel={1 - SPD},
                xlabel={AUROC},
                xmin=0.5,
                ymax=1,
                grid=both,
                major grid style={line width=0.1pt, draw=gray!50, opacity=0.3},
                minor grid style={line width=0.1pt, draw=gray!30, opacity=0.3},
                ylabel style={yshift=-2pt},
                label style={font=\Large },
                tick label style={font=\large},
                legend style={ 
                    at={(0.5,10.3)}, 
                    anchor=north, 
                    draw=none,
                    legend columns=2,
                    font=\footnotesize, 
                }, 
                legend to name=combinedlegend
            ]

            \addplot[
                scatter src=explicit,
                only marks,
                mark=o,
                mark size=2.0pt,
                color=green,
                table/col sep=comma,
                ]
                table [
                y={1-spds_test}, 
                x={aurocs_test}
            ] {results/avg_curve_Dutch_two_trees.csv};
            \addlegendentry{2TFT (Two trees)}

            \addplot[
                scatter src=explicit,
                only marks,
                mark=triangle,
                mark size=2.0pt,
                color=red,
                table/col sep=comma,
                ]
                table [
                y={1-spds_test}, 
                x={aurocs_test}
            ] {results/avg_curve_Dutch_weighted_combi.csv};
            \addlegendentry{One tree (combined split criterion)}

            \addplot[
                scatter src=explicit,
                only marks,
                mark=square,
                mark size=2.0pt,
                color=blue,
                table/col sep=comma,
                ]
                table [
                y={1-spds_test}, 
                x={aurocs_test}
            ] {results/avg_curve_Dutch_constrained.csv};
            \addlegendentry{One tree (constrained split criterion)}

            \addplot[
                scatter src=explicit,
                only marks,
                mark=diamond,
                mark size=2.0pt,
                color=brown,
                table/col sep=comma,
                ]
                table [
                y={1-spds_test}, 
                x={aurocs_test}
            ] {results/avg_curve_Dutch_backtracking.csv};
            \addlegendentry{DTFC (backtracking)}

            \addplot[
                scatter src=explicit,
                only marks,
                mark=+,
                mark size=2pt,
                color=darkgray,
                table/col sep=comma,
                ]
                table [
                y={1-spds_test}, 
                x={aurocs_test}
            ] {results/avg_curve_Dutch_postprocessing.csv};
            \addlegendentry{Threshold Optimizer}
            
            \end{axis}
        \end{tikzpicture}
        \subcaption{Dutch Census}
    \end{minipage}
    
    \caption{We compare the trade-off curves of the described methods on multiple datasets. The 15 best curves were averaged for each dataset to create the shown curves.}
    \label{fig:trade-offs}
\end{figure*}
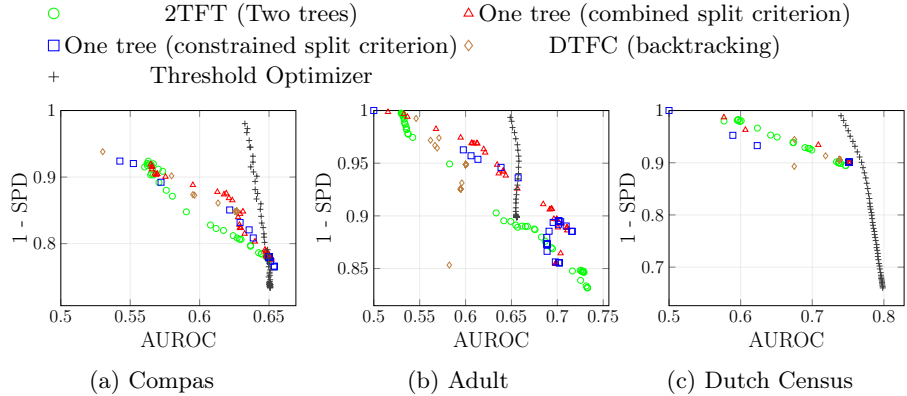

The trade-off curves illustrated in Figure \ref{fig:trade-offs} and the evaluation criteria listed in Table~\ref{tab:results_holdout} demonstrate how the performance of each tree variant evolves under different fairness constraints, i.e., increasing $\gamma$ stepwise in the defined range. The curves are computed by averaging the 15 best curves obtained from the described hold-out procedure.
\begin{table*}[t]
\caption{Evaluation of the trade-off curves for different methods and splitting criteria (SP). For a detailed description of the evaluation metrics we chose, we refer to Section \ref{sec:definitions}. We print in boldface the best  value of each metric for each dataset, where the baseline is not considered. The averaged AUTOC value is marked with~* if there was a statistical significance.}
\centering
\setlength{\tabcolsep}{1mm}
\resizebox{0.99\textwidth}{!}{%
\begin{tabular}{ll|ccccc} 
\hline
Dataset & Method & AUTOC 
& \parbox{1.8cm}{\vspace{2pt}\centering Number of \newline local \newline Pareto points\vspace{2pt}} 
& \parbox{1.8cm}{\vspace{2pt}\centering Number of \newline unique \newline points \vspace{2pt}} 
& \parbox{1.8cm}{\vspace{2pt}\centering Number of \newline unique \newline Pareto points\vspace{2pt}} 
& \parbox{1.8cm}{\vspace{2pt}\centering Variance of \newline pairwise \newline distances\vspace{2pt}} \\ \hline

Adult & 2TFT (Two trees) & 0.434 $\pm$ 0.027 & 39.3 $\pm$ 9.7 & $\textbf{23.7}^* \pm$ 8.4 & \textbf{7.3} $\pm$ 8.4 & $\textbf{0.0095}^* \pm$ 0.0056 \\
 & One tree (combined SP) & \textbf{0.451} $\pm$ 0.005 & 25.2 $\pm$ 9.4 & 10.7 $\pm$ 4.1 & 3.3 $\pm$ 2.1 & 0.0041 $\pm$ 0.0013 \\
 & One tree (constrained SP) & 0.448 $\pm$ 0.005 & 29.7 $\pm$ 9.5 & 8.1 $\pm$ 1.1 & 2.1 $\pm$ 0.9 & 0.0021 $\pm$ 0.0012 \\
 & DTFC (backtracking) & 0.220 $\pm$ 0.207 & \textbf{42.9} $\pm$ 6.6 & 3.9 $\pm$ 2.7 & 0.8 $\pm$ 0.8 & 0.0015 $\pm$ 0.0016 \\
 & Threshold optimizer & 0.339 $\pm$ 0.205 & 19.0 $\pm$ 18.9 & 28.5 $\pm$ 17.8 & 22.6 $\pm$ 14.6 & 0.0014 $\pm$ 0.0010 \\ \hline
COMPAS & 2TFT (Two trees) & 0.402 $\pm$ 0.011 & 34.5 $\pm$ 10.3 & $\textbf{12.5}^* \pm$ 4.3 & $\textbf{4.8} \pm$ 3.1 & $\textbf{0.0062}^* \pm$ 0.0034 \\
 & One tree (combined SP) & 0.402 $\pm$ 0.009 & 35.5 $\pm$ 11.8 & 8.0 $\pm$ 1.3 & 3.2 $\pm$ 1.2 & 0.0036 $\pm$ 0.0017 \\
 & One tree (constrained SP) & 0.400 $\pm$ 0.008 & 39.1 $\pm$ 17.4 & 7.9 $\pm$ 1.1 & 2.3 $\pm$ 1.4 & 0.0031 $\pm$ 0.0011 \\
 & DTFC (backtracking) & $\textbf{0.430}^* \pm$ 0.005 & $\textbf{48.7}^* \pm$ 1.1 & 5.1 $\pm$ 0.6 & 1.1 $\pm$ 0.9 & 0.0009 $\pm$ 0.0003 \\
 & Threshold optimizer & 0.406 $\pm$ 0.018 & 10.5 $\pm$ 4.9 & 36.1 $\pm$ 5.6 & 25.7 $\pm$ 7.0 & 0.0056 $\pm$ 0.0014 \\ \hline
Dutch & 2TFT (Two trees) & 0.463 $\pm$ 0.004 & 46.5 $\pm$ 4.9 & $\textbf{4.3}^* \pm$ 0.9 & \textbf{0.5} $\pm$ 0.7 & \textbf{0.0053} $\pm$ 0.0023 \\
 & One tree (combined SP) & \textbf{0.464} $\pm$ 0.004 & 49.8 $\pm$ 0.4 & 2.9 $\pm$ 0.6 & 0.2 $\pm$ 0.4 & 0.0025 $\pm$ 0.0002 \\
 & One tree (constrained SP) & 0.459 $\pm$ 0.003 & \textbf{50.0} $\pm$ 0.0 & 3.0 $\pm$ 0.0 & 0.0 $\pm$ 0.0 & 0.0020 $\pm$ 0.0007 \\
 & DTFC (backtracking) & 0.455 $\pm$ 0.010 & 49.5 $\pm$ 0.5 & 2.1 $\pm$ 1.0 & \textbf{0.5} $\pm$ 0.5 & 0.0008 $\pm$ 0.0011 \\
 & Threshold optimizer & 0.419 $\pm$ 0.034 & 30.3 $\pm$ 8.1 & 47.7 $\pm$ 4.6 & 18.3 $\pm$ 7.8 & 0.0090 $\pm$ 0.0049 \\ \hline
\end{tabular}
\label{tab:results_holdout}
}
\end{table*}
Table \ref{tab:results_holdout} gives the measures used in the evaluation process, computed on the optimized curves.
For each dataset, we test whether the method with the highest value is significantly better using a pairwise two-sided t-test with $p<0.05$. This verifies whether the 15 found values of one method differ statistically from those of another. If so, we mark the value with~*. For this test, we do not consider the baseline. It serves as a visual comparison, although its AUTOC values might be lower due to the steepness of the curves.

In terms of the AUTOC measure, we notice that the one-tree approach with the combined weighted objective function achieves the best value for two of the datasets (Adult and Dutch), but the difference is not statistically significant. 2TFT, the approach with two objective functions (i.e., with trees $T_y$ and $T_s$), shows a comparable AUTOC value on Dutch Census, with the combined objective function being only slightly better.
The backtracking variation DTFC is significantly better on the COMPAS dataset in terms of AUTOC, even though fewer combinations of hyperparameters were tested.
The number of local Pareto points, however, reaches the maximum for Adult and COMPAS when having one constrained objective extended with the backtracking variation. Only for Dutch, the values are rather similar across all methods, with the constrained objective (without backtracking) being slightly better than the others.

Regarding the number of unique points per curve, we obtain substantially different results.
2TFT consistently performs best, making it the most suitable method for producing diverse solutions across the trade-off space and enabling finer granularity in decision-making, as seen in Figure~\ref{fig:trade-offs}.
The combined objective function for one tree can be identified as the second-best. The same pattern holds for distribution variance, indicating that the two-tree method yields a more evenly distributed trade-off curve and explores the trade-off space more extensively, approaching even the baseline.
The number of unique Pareto points — defined as the intersection of Pareto-optimal and unique points — is also highest for the two-tree method, which shares the lead with DTFC on the Dutch dataset. However, we note that most of the methods show a very low average number of unique Pareto points.

The results suggest that having one objective may be more suitable when aiming for a smaller set of strong outcomes, although we can identify regions in the trade-off space in Figure \ref{fig:trade-offs}, where points of the dual-approach dominate. In contrast, if a broad variety of trade-off solutions is desired, having two objectives, separated into two trees, may be more appropriate.
Additionally, in Figure \ref{fig:trade-offs}, the backtracking variant appears to fill the gaps in the trade-off curve of the constrained approach without backtracking, even though fewer hyperparameters were tested due to runtime limitations.

The evaluation approach used in this study has certain limitations. One key challenge is the difficulty of reducing an entire performance curve to a single numerical value, as this simplification may not fully capture the complexities present in the trade-off process. 
Further, the AUTOC metric, while useful for quantitative comparison, may not always align with human judgment in selecting the most suitable trade-off curve. 

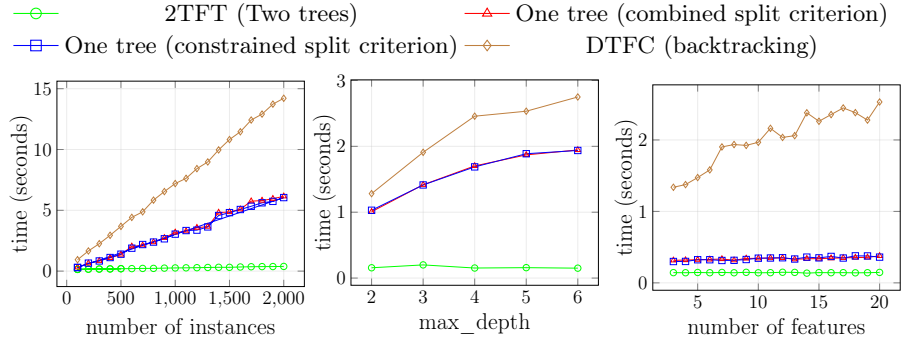
\begin{figure*}[t]
    \centering
    \ref{timeslegend}
    
    \begin{minipage}[t]{0.32\textwidth}
        \centering
        \begin{tikzpicture}[scale=0.6]
            \begin{axis}[
                scale=0.8, 
                ylabel={time (seconds)},
                xlabel={number of instances},
                grid=both,
                major grid style={line width=0.1pt, draw=gray!50, opacity=0.3},
                minor grid style={line width=0.1pt, draw=gray!30, opacity=0.3},
                ylabel style={yshift=-6pt},
                label style={font=\Large },
                tick label style={font=\large},
                legend style={ 
                    at={(0.5,10.3)}, 
                    anchor=north, 
                    draw=none,
                    legend columns=2,
                    font=\footnotesize, 
                }, 
                legend to name=timeslegend
            ]

            \addplot[
                scatter src=explicit,
                mark=o,
                mark size=2.0pt,
                color=green,
                table/col sep=comma,
                ]
                table [
                y={time elapsed}, 
                x={number of rows}
            ] {results_time_inst/time_Adult_predict_proba_FAIRfairness_gain_PERFsklearn_gain_s.csv};
            \addlegendentry{2TFT (Two trees)}

            \addplot[
                scatter src=explicit,
                mark=triangle,
                mark size=2.0pt,
                color=red,
                table/col sep=comma,
                ]
                table [
                y={time elapsed}, 
                x={number of rows}
            ] {results_time_inst/time_Adult_predict_proba_weighted_combi_5.csv};
            \addlegendentry{One tree (combined split criterion)}

            \addplot[
                scatter src=explicit,
                mark=square,
                mark size=2.0pt,
                color=blue,
                table/col sep=comma,
                ]
                table [
                y={time elapsed}, 
                x={number of rows}
            ] {results_time_inst/time_Adult_predict_proba_threshold_gain_s_5.csv};
            \addlegendentry{One tree (constrained split criterion)}

            \addplot[
                scatter src=explicit,
                mark=diamond,
                mark size=2.0pt,
                color=brown,
                table/col sep=comma,
                ]
                table [
                y={time elapsed}, 
                x={number of rows}
            ] {results_time_inst/time_Adult_predict_proba_backtracking_5.csv};
            \addlegendentry{DTFC (backtracking)}

            \end{axis}
        \end{tikzpicture}
        \end{minipage}
        \begin{minipage}[t]{0.32\textwidth}
        \centering
        \begin{tikzpicture}[scale=0.6]
            \begin{axis}[
                scale=0.8, 
                ylabel={time (seconds)},
                xlabel={max\_depth},
                grid=both,
                major grid style={line width=0.1pt, draw=gray!50, opacity=0.3},
                minor grid style={line width=0.1pt, draw=gray!30, opacity=0.3},
                ylabel style={yshift=-6pt},
                label style={font=\Large },
                tick label style={font=\large},
                legend style={ 
                    at={(0.5,10.3)}, 
                    anchor=north, 
                    draw=none,
                    legend columns=2,
                    font=\footnotesize, 
                }, 
                legend to name=timeslegend
            ]

            \addplot[
                scatter src=explicit,
                mark=o,
                mark size=2.0pt,
                color=green,
                table/col sep=comma,
                ]
                table [
                y={time elapsed}, 
                x={max_depth}
            ] {results_time_depth/time_Adult_predict_proba_FAIRfairness_gain_PERFsklearn_gain_s.csv};
            \addlegendentry{2TFT (Two trees)}

            \addplot[
                scatter src=explicit,
                mark=triangle,
                mark size=2.0pt,
                color=red,
                table/col sep=comma,
                ]
                table [
                y={time elapsed}, 
                x={max depth}
            ] {results_time_depth/time_Adult_predict_proba_weighted_combi_5.csv};
            \addlegendentry{One tree (combined split criterion)}

            \addplot[
                scatter src=explicit,
                mark=square,
                mark size=2.0pt,
                color=blue,
                table/col sep=comma,
                ]
                table [
                y={time elapsed}, 
                x={max depth}
            ] {results_time_depth/time_Adult_predict_proba_threshold_gain_s_5.csv};
            \addlegendentry{One tree (constrained split criterion)}

            \addplot[
                scatter src=explicit,
                mark=diamond,
                mark size=2.0pt,
                color=brown,
                table/col sep=comma,
                ]
                table [
                y={time elapsed}, 
                x={max depth}
            ] {results_time_depth/time_Adult_predict_proba_backtracking_5.csv};
            \addlegendentry{DTFC (backtracking)}

            \end{axis}
        \end{tikzpicture}
        \end{minipage}
        \begin{minipage}[t]{0.32\textwidth}
        \centering
        \begin{tikzpicture}[scale=0.6]
            \begin{axis}[
                scale=0.8, 
                ylabel={time (seconds)},
                xlabel={number of features},
                grid=both,
                major grid style={line width=0.1pt, draw=gray!50, opacity=0.3},
                minor grid style={line width=0.1pt, draw=gray!30, opacity=0.3},
                ylabel style={yshift=-6pt},
                label style={font=\Large},
                tick label style={font=\large},
                legend style={ 
                    at={(0.5,10.3)}, 
                    anchor=north, 
                    draw=none,
                    legend columns=2,
                    font=\footnotesize, 
                }, 
                legend to name=timeslegend,
            ]

            \addplot[
                scatter src=explicit,
                mark=o,
                mark size=2.0pt,
                color=green,
                table/col sep=comma,
                ]
                table [
                y={time elapsed}, 
                x={number of cols}
            ] {results_time_feat/time_Adult_predict_proba_FAIRfairness_gain_PERFsklearn_gain_s.csv};
            \addlegendentry{2TFT (Two trees)}

            \addplot[
                scatter src=explicit,
                mark=triangle,
                mark size=2.0pt,
                color=red,
                table/col sep=comma,
                ]
                table [
                y={time elapsed}, 
                x={number of columns}
            ] {results_time_feat/time_Adult_predict_proba_weighted_combi_5.csv};
            \addlegendentry{One tree (combined split criterion)}

            \addplot[
                scatter src=explicit,
                mark=square,
                mark size=2.0pt,
                color=blue,
                table/col sep=comma,
                ]
                table [
                y={time elapsed}, 
                x={number of columns}
            ] {results_time_feat/time_Adult_predict_proba_threshold_gain_s_5.csv};
            \addlegendentry{One tree (constrained split criterion)}

            \addplot[
                scatter src=explicit,
                mark=diamond,
                mark size=2.0pt,
                color=brown,
                table/col sep=comma,
                ]
                table [
                y={time elapsed}, 
                x={number of columns}
            ] {results_time_feat/time_Adult_predict_proba_backtracking_5.csv};
            \addlegendentry{DTFC (backtracking)}

            \end{axis}
        \end{tikzpicture}
        \end{minipage}
    \caption{Runtime analysis on Adult. Each run generates a trade-off curve using 100 instances, depth 3, and 3 features, increasing one of the parameters stepwise.}
    \label{fig:times}
\end{figure*}

Figure \ref{fig:times} analyzes running times depending on the number of instances, maximum depth of the tree, and number of features in the dataset. In each run, one trade-off curve is created using 100 instances, a maximum depth of 3, and 3
features, where one of the parameters is increased stepwise.
2TFT runs much faster than one-tree variants, as it builds only two trees reused across the curve. In contrast, one-tree models retrain a tree for each $\gamma$ value. Among them, the backtracking approach is the slowest due to its extensive search. If performance differences are minimal, the dual-tree variant may be the more practical choice.


From an interpretability perspective, the dual-tree approach offers advantages over the single-tree approach. In the dual-tree method, understanding the trade-off curve requires analyzing only two trees, regardless of the number of trade-off parameter values of $\gamma$ used to construct the curve. 
Consequently, the complexity remains constant even when the trade-off space expands.
In contrast, the single-tree method produces a separate tree for each $\gamma$, complicating interpretation as the trade-off space grows.
The approach with two objective functions provides additional transparency by allowing an examination of the predictions separately from the individual trees. This enables a clear understanding of how the model would behave if it was optimized solely for performance or fairness, facilitating a direct comparison of decision paths and decisions between the two objectives. In contrast, the single-tree method lacks this separation, offering only a single set of predictions and decision paths.

\section{Conclusion}
\label{sec:conclusion}

In this study, we revisited fair tree learning and compared methods for balancing fairness and performance in decision trees across multiple datasets. We examined three different approaches for designing fair tree learning algorithms: (i) one combined objective function in $y$, $s$ and $T$, (ii) one objective function in $y$ and $T$ with a constraint in $s$ and $T$, and (iii) two objective functions, the first being a function only in $y$ and $T$, and the second a function in $s$ and $T$, resulting in two separate trees.
We introduced a backtracking variant which can be classified into (ii) and an approach for (iii). 
Methods designed according to (i) or (ii) yielded smaller unique sets of strong outcomes, and the backtracking variant consistently achieved a high number of local Pareto points. The method from design variant (iii) provided broader and more diverse trade-off curves.
The two-tree method also offers interpretability advantages, requiring only two trees to understand the full trade-off, and allowing explicit comparison between performance- and fairness-driven decisions.

In future work, it would be intriguing to study, amongst others, the behavior of optimal decision trees in this space and non-linear combinations of models.

\subsubsection{\ackname}
The research in this paper was supported by the “TOPML: Trading Off Non-Functional Properties of Machine Learning” project funded by Carl Zeiss Foundation, grant number P2021-02-014.

%
%
%
\bibliographystyle{splncs04}
\bibliography{references}

\end{document}